\documentclass[sigconf]{acmart}
\AtBeginDocument{%
  }

\setcopyright{acmlicensed}
\copyrightyear{2026}
\acmYear{2026}
\acmDOI{XXXXXXX.XXXXXXX}
\acmConference[EMF Workshop, HRI 2026]{}{March 16,
  2026}{Edinburgh, Scotland}
\acmBooktitle{EMF Workshop, HRI 2026, Edinburgh, Scotland}
\acmISBN{978-1-4503-XXXX-X/2018/06}

\begin{document}

%\title{The Need for Bidirectional Streaming in HRI}
\title{Robots That Talk Too Much and Don't Listen Enough}

\author{Graham Wilcock}
\affiliation{%
  \institution{CDM Interact and University of Helsinki}
  \city{Helsinki}
  \country{Finland}}
% \affiliation{%
%   \institution{University of Helsinki}
%   \city{Helsinki}
%   \country{Finland}}
\email{graham.wilcock@cdminteract.com}

\author{Kristiina Jokinen}
\authornote{Also Visiting Professor at Institute of Science Tokyo, Tokyo, Japan.}
\affiliation{%
  \institution{University of Helsinki and CNRS-AIST Joint Robotics Lab} 
  %National Institute of Advanced Industrial Science and Technology}
  \city{Tsukuba}
  \country{Japan}}
%   \institution{University of Helsinki and Institute of Science Tokyo}
%   \city{Tokyo}
%   \country{Japan}}
% \affiliation{%
%   \institution{University of Helsinki and Institute of Science Tokyo}
%   \city{Helsinki}
%   \country{Finland}}
\email{kristiina.jokinen@helsinki.fi}

\renewcommand{\shortauthors}{G. Wilcock and K. Jokinen}

\begin{abstract}
Until recently, the biggest obstacles to natural spoken 
interaction with social robots were speech recognition errors 
and turn-taking failures.
Advances in speech recognition have enabled much better spoken interaction, 
but turn-taking failures and other misunderstandings
continue to be far too prevalent.
This short paper looks back at earlier methods for
improving speech recognition and turn-taking in interaction 
with social robots, and then looks at what can be done today,
arguing that multimodal bidirectional streaming 
is needed to reduce the frequency of turn-taking failures.
\end{abstract}





\maketitle

\section{Introduction}

% A long time ago, robots could not talk very much. 
% They would say \textit{Moving left 10cm} or something equally robotic.
% Now we have social robots that can talk a lot. 
Once upon a time, robots did not talk much. 
They would say \textit{Moving left 10cm} or something equally robotic.
Then they learned to talk a lot. 
We made our own contribution to this by developing WikiTalk for Nao robots in 2012 \cite{Csapo:ea:CogInfoCom:12},
which is described in Section~\ref{sec:past}.

Unfortunately, many social robots now have a problem.
They talk a lot, but they don't listen enough.
For example, many shopping malls in Japan introduced 
humanoid robots about 10 years ago 
to welcome customers and provide information. 
However, the robots had problems understanding what the customers asked, 
and they kept saying \textit{Irrashaimase!} (Welcome) 
repeatedly without answering customers' questions.
It has been said that by 2019, a majority of Pepper robots used in Japan 
were returned without renewing their rental 
contracts.
\textit{"Only about 15\% of businesses renewed their three-year contracts. 
The robot's limited conversational ability, inability to handle complex tasks, 
and high ongoing costs made it difficult to justify for many 
organizations."} \cite{Robozaps:Pepper:26}.

% Unfortunately, many social robots have a problem.
% They talk a lot, but they don't listen enough.
% For example, many shopping malls in Japan introduced 
% humanoid robots about 10 years ago 
% to welcome customers and provide information. 
% However, the robots had problems understanding what the customers asked, 
% and they kept saying \textit{Irrashaimase!} (Welcome) 
% repeatedly without answering customers' questions.
% It is said that by 2019, a majority of Pepper robots used in Japan 
% were returned without renewing their rental 
% contracts\footnote{\textit{"Only about 15\% of businesses renewed their three-year contracts. 
% The robot's limited conversational ability, inability to handle complex tasks, 
% and high ongoing costs made it difficult to justify for many 
% organizations."} \cite{Robozaps:Pepper:26}}.

Until recently, the biggest obstacles to natural spoken 
interaction with social robots were speech recognition errors 
and turn-taking failures.
Impressive advances in speech recognition have enabled much better spoken interaction, 
but turn-taking failures and other misunderstandings
continue to be far too prevalent \cite{Skantze:21}.

This paper looks back at some earlier methods for
improving speech recognition and turn-taking in interaction 
with social robots, and then looks at what can be done today.
The paper argues that bidirectional streaming 
is urgently needed to reduce the frequency of turn-taking failures.
Bidirectional audio streaming allows robots to listen 
while they are speaking, in case the user attempts to speak.
Video streaming of user facial expressions allows robots to 
detect, while they are speaking, that the interaction is not going smoothly
or something is wrong.
Both improvements will enable more natural and fluid interactions between humans and robots.

%\newpage
\section{Earlier times (2012)}
%\section{Past (2012)}
\label{sec:past}

%\newpage
\begin{figure}[htb]
\includegraphics[width=\columnwidth]{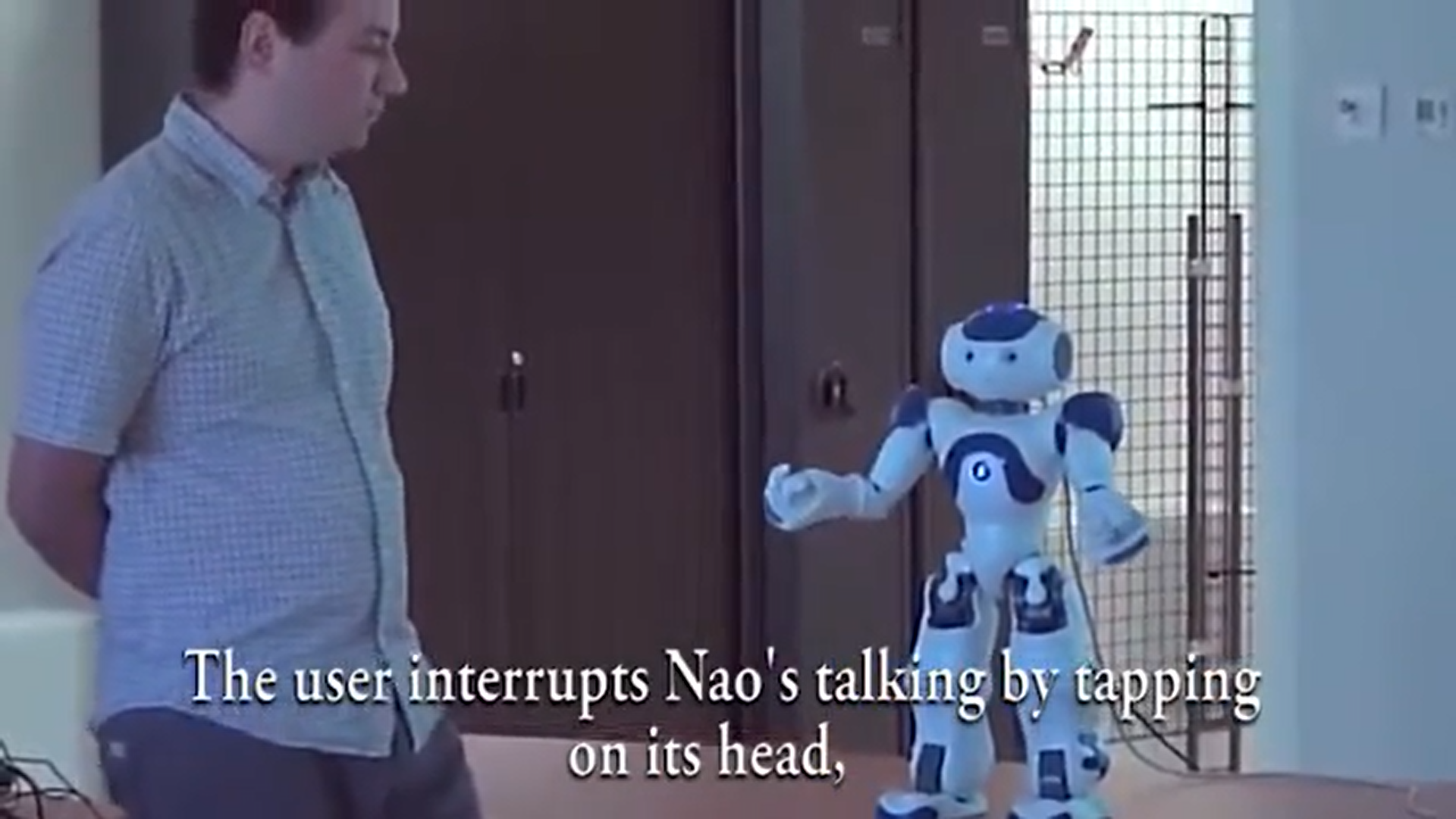}
\caption{Speech recognition and turn-taking with a talkative Nao social robot connected to Wikipedia in 2012.}
\Description{A screenshot from a video showing a dialogue with a Nao robot.}
\label{fig:adam-interrupt}
\end{figure}

An illustration of the earlier challenges of speech recognition 
and turn-taking is seen in the
first demo of WikiTalk \cite{Csapo:ea:CogInfoCom:12},
recorded in 2012 in the 
video\footnote{\url{https://www.youtube.com/watch?v=A7XFcnJE29A}} 
shown in Figure~\ref{fig:adam-interrupt}.
WikiTalk was a spoken dialogue system that enabled Nao robots to talk fluently about thousands of topics
by using extracts from Wikipedia articles. 

The unreliability of speech recognition at the time can be seen 
from the user's behaviour in the video. 
When he speaks, he leans forward to get as close as possible 
to the robot's microphone, as people do when talking 
to someone with a hearing impairment.

Because speech recognition was so unreliable, it helped a lot 
if you could predict what the user would say next. 
The idea was to replace the (then) very difficult problem of open vocabulary 
speech recognition by the much more tractable problem of 
listening for a limited number of predicted words and phrases.

While the robot was talking about a given topic requested by the user,
the system extracted Wikipedia hyperlinks from that topic to related topics,
predicting that the user would probably ask about one of the linked topics next.
When the user mentioned the name of any of these linked topics, the robot could immediately 
make a \textit{smooth topic shift}  
and start talking about the new topic, as described in \cite{Wilcock:QACD:12}. In this way, users could navigate freely around Wikipedia 
following their own personal interests.

This approach of predicting what the user will say next was 
not merely a workaround for the imperfect speech recognition.
It is closer to the way human conversation works. 
People know what their conversation partner is likely to say next, 
and use this knowledge to fill in any words and phrases 
that are not heard clearly.
This is possible because people are aware of the context, 
including the dialogue history, the shared knowledge 
between the partners, 
and the changing environment in which the conversation is taking place.
Even high-quality speech recognition technology makes bad 
mistakes when it has inadequate understanding of the context.

Turn-taking was also a problem, as the Nao robots did not listen 
while they were speaking. 
One approach would be for the robot to pause every few seconds while it is speaking,
to check if the user wants to say something. 
But this would take too long when the robot is reading out
paragraphs from Wikipedia articles, 
because users want to hear whole paragraphs without interruptions.

WikiTalk had a simple solution to this problem. 
The robot would only pause at the end of each paragraph,
but the user could easily interrupt the robot at any time by 
tapping a button on the robot's head.
The robot immediately stops talking, looks up in surprise, and says 
\textit{Oh, sorry! What topic would you like to hear about?}

This convenient interruption mechanism is shown in the video.
It was particularly important if the robot misunderstood or misheard
what the user said, and started talking about the wrong topic.
The user could also use tap the button to temporarily 
stop the robot talking for other reasons, for example to answer a phone call, 
and then say \textit{Continue} to tell the robot to continue where it left off.
In the video the user does this near the beginning.

\section{Recent times (2025)}
%\section{Present (2025)}
\label{sec:present}

\begin{figure}[htb]
\includegraphics[width=\columnwidth]{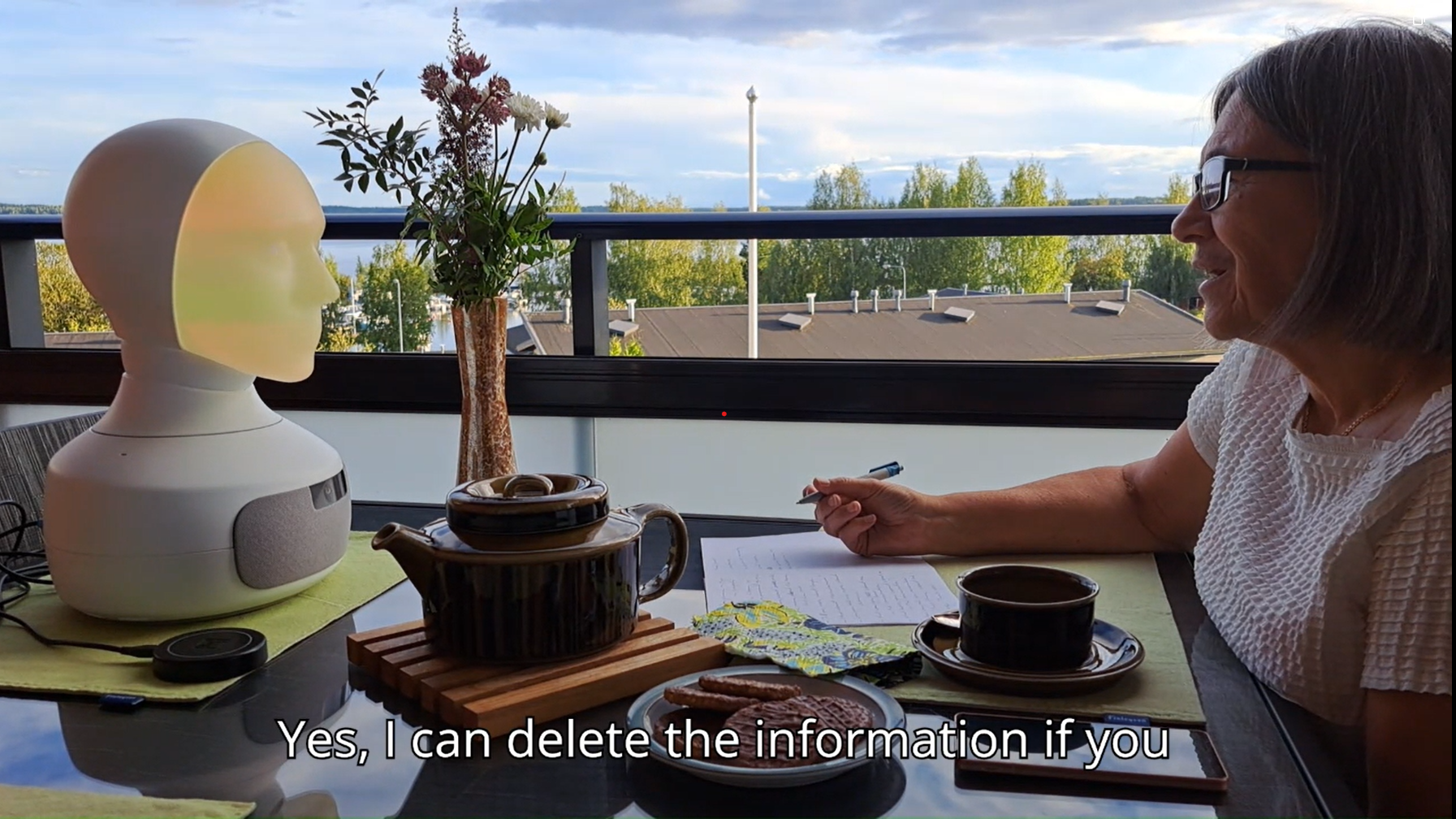}
\caption{Speech recognition and turn-taking with a Furhat robot, 
 a LangGraph AI agent and an OpenAI LLM in 2025.}
\Description{A screenshot from a video showing a dialogue with a Furhat robot.}
\label{fig:delete-memories}
\end{figure}

\begin{table}[tb]
    \small
    \begin{tabular}{| p{0.07\columnwidth} p{0.83\columnwidth} |}
      \hline
    Robot: & \textit{Welcome to LangGraph agent. How can I help you?} \\ 
    User:  & I'm Kristiina, I'm planning a trip to Naples. \\
    Robot: & \textit{Hi, Kristiina, that sounds exciting. What do you have planned for your trip to Naples?} \\ 
    User:  & Well, I'm interested in history, so what would you recommend?\\
    Robot: & \textit{For history, definitely visit the National Archaeological Museum and the ruins of Pompeii. The Castel Nuovo and the catacombs of San Gennaro are also fascinating.} \\
    User:  & Tell me more about Pompeii. \\
    Robot: & \textit{Pompeii is an ancient Roman city that was buried under volcanic ash from Mount Vesuvius in 79 AD.
It's incredibly well preserved, offering a glimpse into daily life in ancient Rome. You can explore homes, streets, and even frescoes.
The site is vast, so consider taking a guided tour to get the most out of your visit.} \\
    User:  & OK. Thank you. \\
    Robot: & \textit{You're welcome, if you have any more questions or need more tips, feel free to ask.} \\
    User:  & You also mentioned Castel Nuovo. Can you tell me more about it?\\
    Robot: & \textit{Sure! Castel Nuovo, also known as Maschio Angioino, is a medieval castle in Naples.
It was built in the 13th century and served as a royal residence. The architecture is impressive, featuring a mix of styles,
and the entrance has a beautiful triumphal arch. Inside, you can explore the royal apartments. Enjoy stunning views of the city from the ramparts.} \\
    User:  & How do I get to to Castel Nuovo? \\
    Robot: & \textit{To get to Castel Nuovo, you can take the metro to the Municipio station, which is just a short walk away. 
Alternatively, if you're in the city centre, it's easily accessible by foot. Just follow signs toward the waterfront and you'll find it near Piazza Municipio.} \\ 
    User:  & OK. Thank you. \\
    Robot: & \textit{You're welcome. Enjoy your trip to Naples. If you have more questions just let me know.} \\
         \hline
    \end{tabular}
    \caption{A dialogue about planning a trip to Naples.}
    \label{tab:planning-trip-dialogue}
\end{table}
  
\begin{table}[tb]
    \small
    \begin{tabular}{| p{0.07\columnwidth} p{0.83\columnwidth} |}
      \hline
    User:  & What have you learned about me in this conversation?
\\
    Robot: & \textit{I've learned that you're Kristiina,
planning a trip to Naples, 
and you're particularly interested in historical sites like Pompeii and Castel Nuovo.
If there's anything else you'd like to share or ask, feel free.} \\ 
    User:  & How long will you store this information?
 \\
    Robot: & \textit{I'll remember this information for our future conversations so I can help you better.
    If you ever want to update or remove anything, just let me know.} \\
    User:  & Alright, if I want to delete some information, can you do it?\\
    Robot: & \textit{If you have any more questions or need assistance, just ask. Enjoy your planning.} \\
    User:  & Can you also delete your information if I want to?\\
    Robot: & \textit{Yes, I can delete the information if you request it. Just let me know whenever you want to do that.} \\
    User:  & Can you delete the information now? \\
    Robot: & \textit{Sure, I'll delete the information now. If you need anything else just let me know, OK?} \\
    User:  & What do you know about me now? \\
    Robot: & \textit{All done. If you have any other questions or need assistance, feel free to ask.} \\ 
    User:  & OK, great. Thank you very much. Bye.\\
    Robot: & \textit{You're welcome, enjoy your day and happy planning for your trip.
} \\
    User:  & Thanks very much. Goodbye. \\
    Robot: & \textit{You're welcome! Goodbye, and take care!} \\ 
    User:  & Do you remember my name? \\
    Robot: & \textit{I don't have your name stored anymore since you asked me to delete the information. 
    If you'd like to share it again or anything else feel free.} \\ 
         \hline
    \end{tabular}
    \caption{A meta-level dialogue about the robot's memory.}
    \label{tab:meta-level-dialogue}
\end{table}

% The LangGraph agentic memory framework can be used in interactions with social robots.
% This section shows how episodic memories of the conversation, that have been saved 
% in long-term memory, can be deleted (forgotten) on request by the user.
Today we have excellent speech recognition technology, 
but there are still many turn-taking failures and other misunderstandings.
For example, Figure~\ref{fig:delete-memories} is from a
video\footnote{\url{https://www.youtube.com/watch?v=f81Xad80AOQ}} 
of a dialogue described by \cite{Wilcock:Jokinen:ICSR:26}.
The Furhat robot is connected to an OpenAI LLM and a LangGraph AI agent
with agentic memory \cite{LangGraph:Memory-GitHub:25}. 

The user is planning a trip to Naples. 
The dialogue for the first part of the video is transcribed in Table~\ref{tab:planning-trip-dialogue}. 
In this part, the user asks simple questions and the robot does most of the talking.
The robot is able to give good recommendations for places to visit in Naples,
and to give more detailed information about the recommended places when asked.
There are no turn-taking failures in this part of the dialogue,

After receiving numerous recommendations for places to visit in Naples, 
the user switches to a meta-level discussion about the conversation 
history and the robot's memory.
The dialogue for this part of the video is transcribed in Table~\ref{tab:meta-level-dialogue}.
The user wants to know what the robot has learnt about her during the conversation,
how long the robot will store this information, and whether 
the robot can delete this information if she asks it to.

Turn-taking goes smoothly at first.
The user asks what the robot has learnt about her during the conversation. 
The robot says \textit{I've learned that you're Kristiina, 
planning a trip to Naples, and you're particularly interested 
in historical sites like Pompeii and Castel Nuovo}. 
The user further asks \textit{How long will you store this information?} 
and the robot replies \textit{I'll remember this information 
for our future conversations so I can help you better.}

But when the user asks \textit{Alright, if I want to delete some information, can you do it?}
the robot replies \textit{If you have any more questions or need assistance, just ask. Enjoy your planning.}
The robot did not listen to the user's question, and does not answer it.

The user persists and asks again \textit{Can you also delete your information if I want to?}
This time the robot listens and replies \textit{Yes, I can delete the information if you request it. Just let me know whenever you want to do that.}
The user then asks \textit{Can you delete the information now?}
and the robot replies \textit{Sure, I'll delete the information now. If you need anything else just let me know, OK?} 

The user tries to check that the information has been deleted 
by asking \textit{What do you know about me now?}
but the robot does not listen to this question, 
and instead replies \textit{All done. If you have any other questions or need assistance, feel free to ask.}

At this point the user is not sure whether the robot has deleted the information or not,
but she decides to end the conversation by saying \textit{OK, great. Thank you very much. Bye.}

After exchanging farewells, the user makes one final effort to gain some clarity 
by asking \textit{Do you remember my name?} 
and the robot replies \textit{I don't have your name stored anymore 
since you asked me to delete the information. 
If you'd like to share it again or anything else feel free.}
% So it is finally clear that the agent has deleted its memories 
% about the user and her interests as requested.
So the robot finally confirms that it has deleted its memories 
about the user and her interests as requested.

Despite excellent speech recognition and 
the power of advanced agentic frameworks with memory systems,
this dialogue illustrates problems in 
turn-taking that remain very common in human-robot interaction. 
Several times the user speaks but the robot does not listen.
This is a big limitation on natural spoken interaction.
The user is forced to learn to wait until the robot has finished speaking 
before asking the next question.

A comprehensive survey of turn-taking problems in human-robot interaction 
is given by Skantze \cite{Skantze:21}.
In the example dialogue, the problems are mainly due to using
the older Furhat Remote API, which only supports simple synchronous clients for  
sequential turn-taking. Some of the problems can be mitigated by
the new Furhat Realtime API \cite{Furhat:RealtimeAPI:25} which uses websockets
to support asynchronous clients for concurrent or event-driven interactions.
% The new Realtime API is explained by Skantze in a demo video\footnote{\url{https://www.youtube.com/watch?v=WtL7tHFHFJo}}. 

\section{Multimodal bidirectional streaming}
%\section{Future}
\label{sec:future}

\begin{figure}[htb]
\includegraphics[width=\columnwidth]{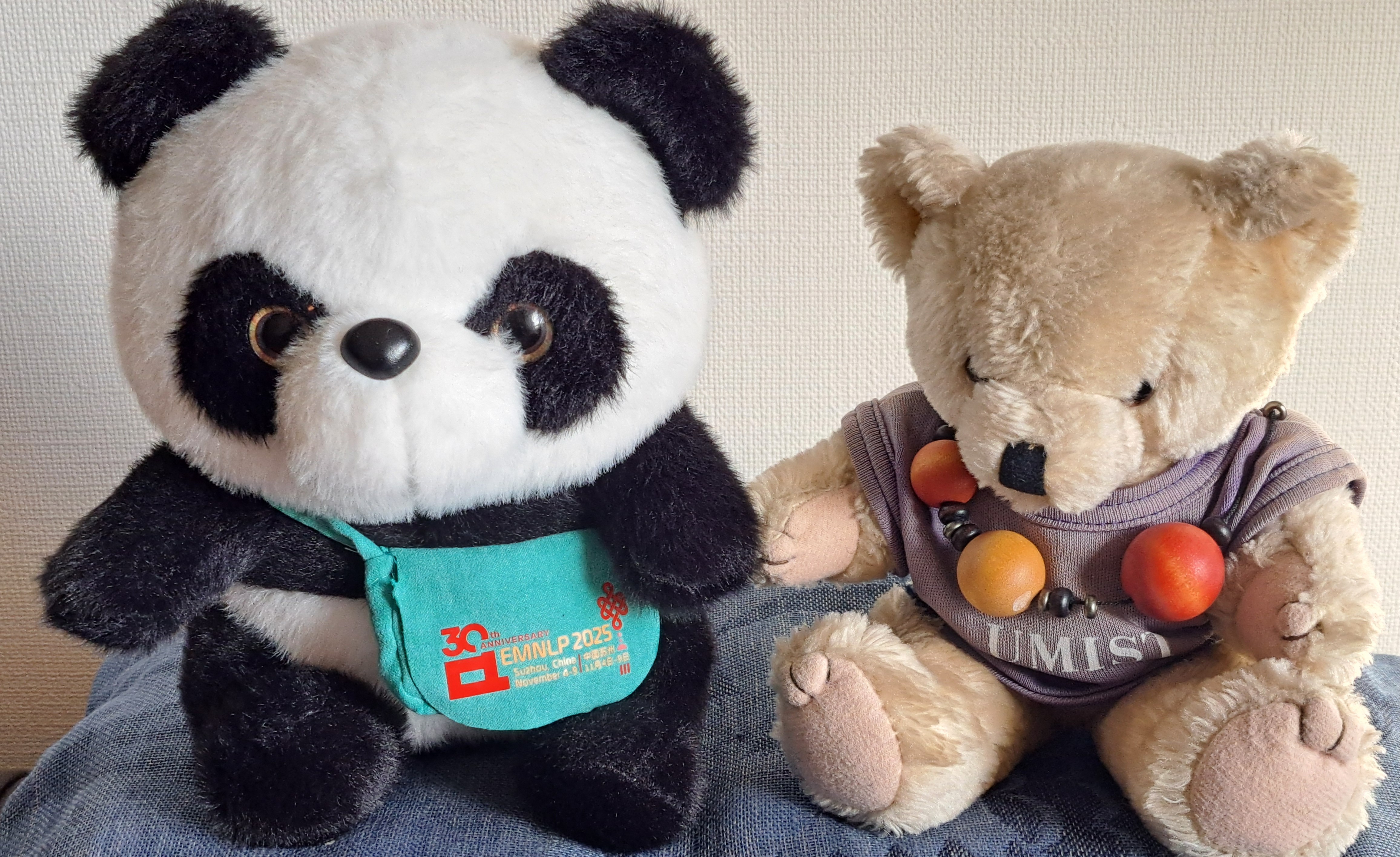}
\caption{Bidirectional audio and video 
streaming for object recognition with Google ADK and a Gemini LLM in 2026.}
\Description{A screenshot from a video showing two teddy bears with acronyms on their chests.}
\label{fig:two-bears}
\end{figure}

In order to reduce the frequency of turn-taking problems of the type seen in Section~\ref{sec:present},
human-robot interaction needs to urgently switch to bidirectional streaming.
The LangGraph agentic framework \cite{LangGraph:Memory-GitHub:25} 
and the Furhat Realtime API \cite{Furhat:RealtimeAPI:25}
mentioned in Section~\ref{sec:present} support asynchronous processing.
We have also experimented with Google ADK (Agent Development Kit) \cite{Google:ADK:26}
which fully supports bidirectional streaming of audio and video using websockets.

Multimodal streaming allows robots to listen and 
observe the scene around them.
Using an ADK agent with video streaming \cite{Google:ADK:ImageVideo:26}, we found that
results in visual object recognition were excellent.
Two of the objects that we used in our experiments were 
the little bears in Figure~\ref{fig:two-bears}.

% Bidirectional audio streaming allows robots to listen 
% while they are speaking, in case the user attempts to speak.
% Using bidirectional audio streaming with an ADK agent \cite{Google:ADK:Bidi:25} 
% on a laptop, 
% we had good results in handling spoken barge-in interruptions.
% Also using an ADK agent with video streaming, we found that
% results in visual object recognition were excellent.
% Two of the objects that we used in our experiments were 
% the little bears in Figure~\ref{fig:two-bears}.

The ADK agent not only recognised the bears, but was also able 
to read the printed acronyms on their chests after some adjustments with the lighting.
At first, the agent thought the smaller bear had a T-shirt with \textit{Feminist} written on it.  
Then it correctly read \textit{UMIST} and used one of its 
tools, Google Search Grounding \cite{Google:ADK:Grounding:26},  
to inform us that \textit{UMIST is University of Manchester Institute of Science and Technology,
which is now merged with University of Manchester}.
It also read \textit{EMNLP 2025} on the panda's bag,
and told us \textit{EMNLP is the Conference on Empirical Methods in Natural Language Processing}.

Video streaming of users' facial expressions can also allow robots to 
detect, while they are speaking, that the interaction is not going smoothly
or something is wrong. 
Instead of the user needing to tap a button on the robot's head 
as in Section~\ref{sec:past} to interrupt the unwanted flow of talking, 
proactive robots can stop talking and ask the user if everything is OK.

% Video streaming of users' facial expressions will also allow robots to 
% detect, while they are speaking, that the interaction is not going smoothly
% or something is wrong. 
% Instead of the user needing to tap a button on the robot's head 
% (as in Section~\ref{sec:past}) to interrupt the unwanted flow of talking, robots will 
% proactively stop talking and ask the user if everything is OK.

Bidirectional audio streaming \cite{Google:ADK:BidiStreaming:26} allows robots to listen 
while they are speaking, so they can hear if the user speaks.
This helps to avoid turn-taking failures of the kind seen in Section~\ref{sec:present}, 
because the robot can stop talking in order to listen
and respond to what the user is saying.
It is especially important when the user pauses briefly and then 
continues speaking to complete what they were saying.
Bidirectional audio streaming with an ADK agent gave good results 
in handling spoken barge-in interruptions by the user.

We plan to use bidirectional audio and video streaming 
with social robots in our ongoing work.
These improvements
%, which are long overdue, 
will surely 
reduce the frequency of turn-taking failures, and will
enable more natural interactions between humans and robots.

% \begin{acks}
% \end{acks}

% \bibliographystyle{ACM-Reference-Format}
% \bibliography{../../bibtex/anthology,../../bibtex/newbib}

\begin{thebibliography}{11}

%%% the Web address.

\ifx \showCODEN    \undefined \def \showCODEN     #1{\unskip}     \fi
\ifx \showISBNx    \undefined \def \showISBNx     #1{\unskip}     \fi
\ifx \showISBNxiii \undefined \def \showISBNxiii  #1{\unskip}     \fi
\ifx \showISSN     \undefined \def \showISSN      #1{\unskip}     \fi
\ifx \showLCCN     \undefined \def \showLCCN      #1{\unskip}     \fi
\ifx \shownote     \undefined \def \shownote      #1{#1}          \fi
\ifx \showarticletitle \undefined \def \showarticletitle #1{#1}   \fi
\ifx \showURL      \undefined \def \showURL       {\relax}        \fi
% invisible to TeX
\providecommand\bibfield[2]{#2}
\providecommand\bibinfo[2]{#2}
\providecommand\natexlab[1]{#1}
\providecommand\showeprint[2][]{arXiv:#2}

\bibitem[Csapo et~al\mbox{.}(2012)]%
        {Csapo:ea:CogInfoCom:12}
\bibfield{author}{\bibinfo{person}{Adam Csapo}, \bibinfo{person}{Emer
  Gilmartin}, \bibinfo{person}{Jonathan Grizou}, \bibinfo{person}{JingGuang
  Han}, \bibinfo{person}{Raveesh Meena}, \bibinfo{person}{Dimitra Anastasiou},
  \bibinfo{person}{Kristiina Jokinen}, {and} \bibinfo{person}{Graham Wilcock}.}
  \bibinfo{year}{2012}\natexlab{}.
\newblock \showarticletitle{Multimodal Conversational Interaction with a
  Humanoid Robot}. In \bibinfo{booktitle}{\emph{Proceedings of 3rd IEEE
  International Conference on Cognitive Infocommunications (CogInfoCom 2012)}}.
  \bibinfo{publisher}{IEEE}, \bibinfo{address}{Kosice, Slovakia},
  \bibinfo{pages}{667--672}.
\newblock


\bibitem[Fankhauser(2026)]%
        {Robozaps:Pepper:26}
\bibfield{author}{\bibinfo{person}{Dean Fankhauser}.}
  \bibinfo{year}{2026}\natexlab{}.
\newblock \bibinfo{title}{SoftBank Pepper Review 2026: Full Specs, Pros, Cons
  and Verdict}.
\newblock
  \bibinfo{howpublished}{\url{https://blog.robozaps.com/b/softbank-pepper-review}}.
\newblock


\bibitem[{Furhat Robotics}(2025)]%
        {Furhat:RealtimeAPI:25}
\bibfield{author}{\bibinfo{person}{{Furhat Robotics}}.}
  \bibinfo{year}{2025}\natexlab{}.
\newblock \bibinfo{title}{{Realtime API}}.
\newblock
  \bibinfo{howpublished}{\url{https://docs.furhat.io/realtime-api/intro}}.
\newblock


\bibitem[Google(2026a)]%
        {Google:ADK:26}
\bibfield{author}{\bibinfo{person}{Google}.} \bibinfo{year}{2026}\natexlab{a}.
\newblock \bibinfo{title}{{Agent Development Kit (ADK)}}.
\newblock \bibinfo{howpublished}{\url{https://adk.dev/}}.
\newblock


\bibitem[Google(2026b)]%
        {Google:ADK:Grounding:26}
\bibfield{author}{\bibinfo{person}{Google}.} \bibinfo{year}{2026}\natexlab{b}.
\newblock \bibinfo{title}{{Google Search Grounding for agents}}.
\newblock
  \bibinfo{howpublished}{\url{https://adk.dev/grounding/google_search_grounding/}}.
\newblock


\bibitem[Google(2026c)]%
        {Google:ADK:ImageVideo:26}
\bibfield{author}{\bibinfo{person}{Google}.} \bibinfo{year}{2026}\natexlab{c}.
\newblock \bibinfo{title}{{How to Use Image and Video}}.
\newblock
  \bibinfo{howpublished}{\url{https://adk.dev/streaming/dev-guide/part5/##how-to-use-image-and-video}}.
\newblock


\bibitem[Google(2026d)]%
        {Google:ADK:BidiStreaming:26}
\bibfield{author}{\bibinfo{person}{Google}.} \bibinfo{year}{2026}\natexlab{d}.
\newblock \bibinfo{title}{{What is Bidi-streaming?}}
\newblock
  \bibinfo{howpublished}{\url{https://adk.dev/streaming/dev-guide/part1/##11-what-is-bidi-streaming}}.
\newblock


\bibitem[LangGraph(2025)]%
        {LangGraph:Memory-GitHub:25}
\bibfield{author}{\bibinfo{person}{LangGraph}.}
  \bibinfo{year}{2025}\natexlab{}.
\newblock \bibinfo{title}{Memory}.
\newblock
  \bibinfo{howpublished}{\url{https://langchain-ai.github.io/langgraph/concepts/memory/}}.
\newblock


\bibitem[Skantze(2021)]%
        {Skantze:21}
\bibfield{author}{\bibinfo{person}{Gabriel Skantze}.}
  \bibinfo{year}{2021}\natexlab{}.
\newblock \showarticletitle{Turn-taking in Conversational Systems and
  Human-Robot Interaction: A Review}.
\newblock \bibinfo{journal}{\emph{Computer Speech \& Language}}
  \bibinfo{volume}{67} (\bibinfo{year}{2021}), \bibinfo{pages}{101178}.
\newblock
\showISSN{0885-2308}
\urldef\tempurl%
\url{https://www.sciencedirect.com/science/article/pii/S088523082030111X}
\showURL{%
\tempurl}


\bibitem[Wilcock(2012)]%
        {Wilcock:QACD:12}
\bibfield{author}{\bibinfo{person}{Graham Wilcock}.}
  \bibinfo{year}{2012}\natexlab{}.
\newblock \showarticletitle{{WikiTalk}: A Spoken {Wikipedia}-based Open-Domain
  Knowledge Access System}. In \bibinfo{booktitle}{\emph{Proceedings of the
  COLING 2012 Workshop on Question Answering for Complex Domains}}.
  \bibinfo{address}{Mumbai, India}, \bibinfo{pages}{57--69}.
\newblock


\bibitem[Wilcock and Jokinen(2026)]%
        {Wilcock:Jokinen:ICSR:26}
\bibfield{author}{\bibinfo{person}{Graham Wilcock} {and}
  \bibinfo{person}{Kristiina Jokinen}.} \bibinfo{year}{2026}\natexlab{}.
\newblock \showarticletitle{Cognitive Architecture and Episodic Memory for
  Conversational Grounding}.
\newblock
\newblock
\shownote{18th International Conference on Social Robotics (ICSR 2026), London,
  UK. To appear in Lecture Notes in Computer Science, LNAI 16772, Springer.}.


\end{thebibliography}



\end{document}